\documentclass[10pt,twocolumn,letterpaper]{article}

\usepackage{cvpr}
\usepackage{times}
\usepackage{epsfig}
\usepackage{graphicx}
\usepackage{amsmath}
\usepackage{amssymb}
\usepackage[ruled]{algorithm2e} 
\usepackage{bm}
\usepackage{multirow}
\usepackage{subfig}
\usepackage{float}
\usepackage{mathrsfs}
\usepackage{array}
\usepackage{esvect}

\usepackage[pagebackref=true,breaklinks=true,letterpaper=true,colorlinks,bookmarks=false]{hyperref}

\cvprfinalcopy 

\ifcvprfinal\fi

\begin{document}

\title{Image-Image Domain Adaptation with Preserved Self-Similarity and Domain-Dissimilarity for Person Re-identification}

\author{Weijian Deng$^{\dag}$, \ Liang Zheng$^{\ddag}$$^{\S}$,  \ Qixiang Ye$^{\dag}$, \ Guoliang Kang$^{\ddag}$, \ Yi Yang$^{\ddag}$, \ Jianbin Jiao$^{\dag}\thanks{Corresponding Author 
}$\\
$^{\dag}$University of Chinese Academy of Sciences \quad $^{\ddag}$University of Technology Sydney\\ \quad $^{\S}$Singapore University of Technology and Design\\
\textsl{{\small \{dengwj16, liangzheng06\}@gmail.com, \{qxye, jiaojb\}@ucas.ac.cn, guoliang.kang@student.uts.edu.au,  yi.yang@uts.edu.au}}\\
 }

\maketitle
\thispagestyle{empty}

\begin{abstract}\label{abstract}
Person re-identification (re-ID) models trained on one domain often fail to generalize well to another. In our attempt, we present a ``learning via translation'' framework. In the baseline, we translate the labeled images from source to target domain in an unsupervised manner. We then train re-ID models with the translated images by supervised methods. Yet, being an essential part of this framework, unsupervised image-image translation suffers from the information loss of source-domain labels during translation.

Our motivation is two-fold. First, for each image, the discriminative cues contained in its ID label should be maintained after translation. Second, given the fact that two domains have entirely different persons, a translated image should be dissimilar to any of the target IDs. To this end, we propose to preserve two types of unsupervised similarities, 1) self-similarity of an image before and after translation, and 2) domain-dissimilarity of a translated source image and a target image. Both constraints are implemented in the similarity preserving generative adversarial network (SPGAN) which consists of an Siamese network and a CycleGAN. Through domain adaptation experiment, we show that images generated by SPGAN are more suitable for domain adaptation and yield consistent and competitive re-ID accuracy on two large-scale datasets.
\end{abstract}

\section{Introduction}\label{Introduction}

This paper considers domain adaptation in person re-ID. The re-ID task aims at searching for the relevant images to the query. In our setting, the source domain is fully annotated, while the target domain does not have ID labels. In the community, domain adaptation of re-ID is gaining increasing popularity, because 1) of the expensive labeling process and 2) 
when models trained on one dataset are directly used on another, the re-ID accuracy drops dramatically \cite{fan17unsupervised} due to \emph{dataset bias} \cite{DBLP:conf/cvpr/TorralbaE11}.
Therefore, supervised, single-domain re-ID methods may be limited in real-world scenarios, where domain-specific labels are not available. 

\begin{figure}[]
\setlength{\abovecaptionskip}{-0.1cm} 
\setlength{\belowcaptionskip}{-0.2cm}
\begin{center}
\includegraphics[width=0.8 \linewidth]{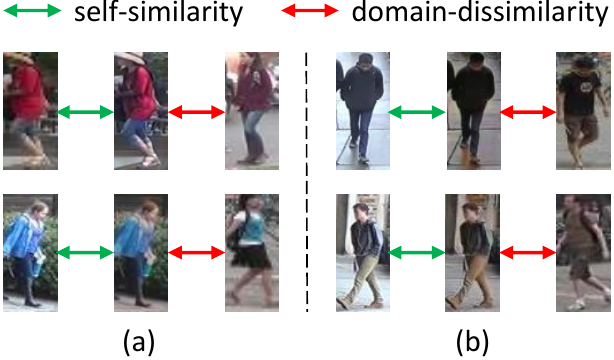}
\end{center}
\caption{Illustration of self-similarity and domain-dissimilarity. In each triplet, left: an source-domain image, middle: a source-target translated version of the source image, right: an arbitrary target-domain image. We require that 1) a source image and its translated image should contain the same ID, \ie, self-similarity, and 2) the translated image should be of a different ID with any target image, \ie, domain dissimilarity. Note: the source and target domains contain entirely different IDs. }
\label{fig:carton}
\end{figure}

\begin{figure*}[t]
\setlength{\abovecaptionskip}{-0.2cm}
\setlength{\belowcaptionskip}{-0.3cm}
\begin{center}
\includegraphics[width=1\linewidth]{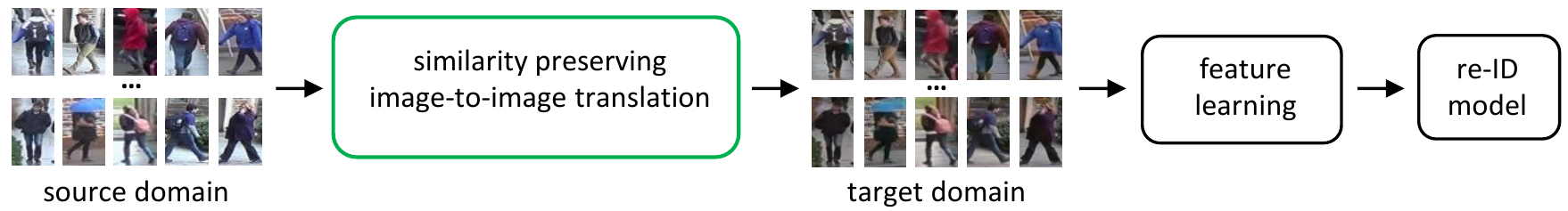}
\end{center}
\caption{Pipeline of the ``learning via translation'' framework. First, we translate the labeled images from a source domain to a target domain in an unsupervised manner. Second, we train re-ID models with the translated images using supervised feature learning methods. The major contribution consists in the first step, \ie,  similarity preserving image-image translation.}
\label{fig:framework}
\end{figure*}

A common strategy for this problem is unsupervised domain adaptation (UDA). But this line of methods assume that the source and target domains contain the same set of classes. Such assumption does not hold for person re-ID because different re-ID datasets usually contain entirely different persons (classes). 
In domain adaptation, a recent trend consists in image-level domain translation \cite{ cycledomain, DBLP:journals/corr/BousmalisSDEK16, DBLP:conf/nips/LiuT16}. \textbf{In the baseline approach}, two steps are involved. First, labeled images from the source domain are transferred to the target domain, so that the transferred image has a similar style with the target. Second, the style-transferred images and their associated labels are used in supervised learning in the target domain. In literature, commonly used style transfer methods include \cite{DBLP:journals/corr/LiuBK17,DiscoGAN, DualGAN, cycle}. In this paper, we use CycleGAN \cite{cycle} following the practice in \cite{DBLP:journals/corr/LiuBK17, cycledomain}.

In person re-ID, there is a distinct yet unconsidered requirement for the baseline described above: the visual content associated with the ID label of an image should be preserved after image-image translation. In our scenario, such visual content usually refers to the underlying (latent) ID information for a foreground pedestrian. To meet this requirement tailored for re-ID, we need additional constraints on the mapping function. 
In this paper, we propose a solution to this requirement, motivated from two aspects. First, a translated image, despite of its style changes, should contain the same underlying identity with its corresponding source image. Second, in re-ID, the source and target domains contain two entirely different sets of identities. Therefore, a translated image should be different from any image in the target dataset in terms of the underlying ID. 

This paper introduces the Similarity Preserving cycle-consistent Generative Adversarial Network (SPGAN), an unsupervised domain adaptation approach which generates images for effective target-domain learning. SPGAN is composed of an Siamese network (SiaNet) and a CycleGAN. Using a contrastive loss, the SiaNet pulls close a translated image and its counter part in the source, and push away the translated image and any image in the target. In this manner, the contrastive loss satisfies the specific requirement in re-ID. Note that, the added constraint is unsupervised, \emph{i.e.,} the source labels are not used during domain adaptation.
During training, in each mini-batch (batch size = 1), a training image is firstly used to update the Generator (of CycleGAN), then the Discriminator (of CycleGAN), and finally the layers in SiaNet. Through the coordination between CycleGAN and SiaNet, we are able to generate samples which not only possess the style of target domain but also preserve their underlying ID information. 

Using SPGAN, we are able to create a dataset on the target domain in an unsupervised manner. The dataset inherits the labels from the source domain and thus can be used in supervised learning in the target domain. The contributions of this work are summarized below:
\begin{itemize}
\item Minor contribution: we present a ``learning via translation'' baseline for domain adaptation in person re-ID.
\item Major contribution: we introduce SPGAN to improve the baseline. SPGAN works by preserving the underlying ID information during image-image translation.
\end{itemize}

\section{Related Work} \label{Related Works}
\textbf{Image-image translation.} 
Image-image translation aims at constructing a mapping function between two domains. A representative method is the conditional GAN \cite{DBLP:journals/corr/IsolaZZE16}, which using paired training data produces impressive transition results. However, the paired training data is often difficult to acquire. Unpaired image-image translation is thus more applicable. To tackle unpaired settings, a cycle consistency loss is introduced by \cite{DiscoGAN, DualGAN, cycle}. In \cite{DBLP:journals/corr/BenaimW17}, an unsupervised distance loss is proposed for one side domain mapping. In \cite{DBLP:journals/corr/LiuBK17}, a general framework is proposed by making a shared latent space assumption. A camera style adaptation method \cite{zhong2018camera} is proposed for re-ID based on CycleGAN.
Our work aims to find a mapping function between the source domain and target domain, and we are more concerned with similarity preserving translation.

Neural style transfer \cite{DBLP:conf/cvpr/GatysEB16, DBLP:conf/eccv/LiW16, DBLP:conf/icml/UlyanovLVL16, DBLP:conf/eccv/JohnsonAF16, DBLP:journals/corr/ChenS16f, DBLP:journals/corr/LiFYWL017, DBLP:journals/corr/HuangB17, DBLP:conf/ijcai/LiWLH17} is another strategy of image-image translation, which aims at replicating the style of one image, while our work focuses on learning the mapping function between two domains, rather than two images. 

\textbf{Unsupervised domain adaptation.}
Our work relates to unsupervised domain adaptation (UDA) where no labeled target images are available during training.
In this community, some methods aim to learn a mapping between source and target distributions \cite{DBLP:conf/eccv/SaenkoKFD10, DBLP:conf/cvpr/GongSSG12,DBLP:conf/iccv/FernandoHST13, DBLP:conf/aaai/SunFS16}.  Correlation Alignment (CORAL) \cite{DBLP:conf/aaai/SunFS16} proposes to match the mean and covariance of two distributions. Recent methods \cite{cycledomain, DBLP:journals/corr/BousmalisSDEK16, DBLP:conf/nips/LiuT16} use an adversarial approach to learn a transformation in the pixel space from one domain to another.
Other methods seek to find a domain-invariant feature space \cite{DBLP:journals/corr/abs-1709-10190, DBLP:conf/cvpr/LongD0SGY13, DBLP:conf/icml/GaninL15, DBLP:conf/icml/LongC0J15, DBLP:journals/corr/TzengHZSD14, DBLP:journals/jmlr/GaninUAGLLML16, DBLP:journals/corr/AjakanGLLM14}. Long \etal \cite{DBLP:conf/icml/LongC0J15} and Tzeng \etal \cite{DBLP:journals/corr/TzengHZSD14} use the Maximum Mean Discrepancy (MMD) \cite{MMD} for this purpose. Ganin \etal \cite{DBLP:journals/jmlr/GaninUAGLLML16} and Ajakan \etal \cite{DBLP:journals/corr/AjakanGLLM14} introduce a domain confusion loss to learn domain-invariant features. 
Different from the settings in this paper, most of the UDA methods assume that class labels  are the same across domains, while different re-ID datasets contain entirely different person identities (classes). Therefore, the approaches mentioned above can not be utilized directly for domain adaptation in re-ID.

\textbf{Unsupervised person re-ID.}
Hand-craft features \cite{DBLP:journals/ivc/MaSJ14,DBLP:conf/eccv/GrayT08,DBLP:conf/cvpr/FarenzenaBPMC10,DBLP:conf/cvpr/MatsukawaOSS16,DBLP:conf/cvpr/LiaoHZL15,DBLP:conf/iccv/ZhengSTWWT15} can be directly employed for unsupervised re-ID. But these feature design methods do not fully exploit rich information from data distribution. 
Some methods are based on saliency statistics \cite{DBLP:conf/cvpr/ZhaoOW13,DBLP:conf/bmvc/WangGX14}. In \cite{CAMEL}, K-means clustering is used for learning an unsupervised asymmetric metric. 
Peng \etal \cite{DBLP:conf/cvpr/PengXWPGHT16} propose an asymmetric multi-task dictionary learning for cross-data transfer.

Recently, several works focus on label estimation of unlabeled target dataset. Ye \etal \cite{DBLP:journals/corr/abs-1709-09297} use graph matching for cross-camera label estimation. Fan \etal \cite{fan17unsupervised} propose a progressive method based on the iterations between K-means clustering and IDE \cite{DBLP:journals/corr/ZhengYH16} fine-tuning. Liu \emph{et al.} \cite{liu2017stepwise} employ a reciprocal search process to refine the estimated labels. Wu \etal \cite{wu2018exploit} propose a dynamic sampling stragy for one-shot video-based re-ID. Our work seeks to learn re-ID models that can be utilized directly to target domain, and can potentially cooperate with label estimation methods in model initialization.
Finally, we would like to refer the reader to the concurrent work named TJ-AIDL \cite{wang2018} that utilizes additional attribute annotation to learn a feature representation space for the unlabeled target dataset.

\section{Proposed Method} \label{Proposed Method}
\subsection{Baseline Overview}\label{sec:baseline}
Given an annotated dataset $\mathcal{S}$ from  source domain and unlabeled dataset $\mathcal{T}$ from  target domain, our goal is to use the labeled source images to train a re-ID model that generalizes well to target domain.  Figure \ref{fig:framework} presents a pipeline of the ``learning via translation'' framework, which consists of two steps, \emph{i.e.,} source-target image translation for training data creation, and supervised feature learning for re-ID.

\begin{itemize}
\setlength{\itemsep}{-0pt}

\item \textbf{Source-target image translation.} Using a generative function $G(\cdot)$ that translates the annotated dataset $\mathcal{S}$ from the source domain to target domain in an unsupervised manner, we ``create'' a labeled training dataset $G(\mathcal{S})$ on the target domain. In this paper, we use CycleGAN \cite{cycle}, following the practice in \cite{DBLP:journals/corr/LiuBK17, cycledomain}.

\item \textbf{Feature learning.} With the translated dataset $G(\mathcal{S})$ that contains labels, feature learning methods are applied to train re-ID models. Specifically, we adopt the same setting as \cite{DBLP:journals/corr/ZhengYH16}, in which the rank-1 accuracy and mAP on the fully-supervised Market-1501 dataset is 75.8\% and 52.2\%.
\end{itemize}

The focus of this paper is to improve Step 1, so that with better training samples, the overall re-ID accuracy can be improved. The experiment will validate the proposed Step 2 ($G_{sp}(\cdot)$) on several feature learning  methods. A brief summary of different methods considered in this paper is presented in Table \ref{table:summary}. We denote the method ``Direct Transfer'' as directly using the training set $\mathcal{S}$ instead of $G(\mathcal{S})$ for model learning. This method yields the lowest accuracy because the style difference between the source and target is not resolved (to be shown in Table \ref{table:cmpbasl}). Using CycleGAN and SPGAN to generate a new training set, which is more style-consistent with the target, yields improvement.

\setlength{\tabcolsep}{4.6pt}
\begin{table}
\setlength{\abovecaptionskip}{-0.1cm} 
\setlength{\belowcaptionskip}{-0.3cm}
\begin{center}
\begin{tabular}{l|ccl}
\hline
Method&Train. Set&Test. Set&Accuracy\\
\hline
\hline
Supervised   & $\mathcal{T}_{train}$ & $\mathcal{T}_{test}$  & +++++\\
Direct Transfer   &  $\mathcal{S}_{train} $ & $\mathcal{T}_{test}$  & ++ \\
CycleGAN (basel.)   & $G(\mathcal{S}_{train}) $ & $\mathcal{T}_{test}$   & +++\\
SPGAN   &  $G_{sp}(\mathcal{S}_{train})$ & $\mathcal{T}_{test}$   & ++++\\
\hline
\end{tabular}
\end{center}
\setlength{\abovecaptionskip}{0cm}
\caption{A brief summary of different methods considered in this paper. ``$G$'' and ``$G_{sp}$'' denote the Generator in CycleGAN and SPGAN, respectively. $\mathcal{S}_{train}$, $\mathcal{T}_{train}$,  $\mathcal{T}_{test}$ denote the training set of the source dataset, the training set and testing set of the target dataset, respectively.}
\label{table:summary}
\end{table}
\subsection{SPGAN: Approach Details}\label{SPGAN}
\subsubsection{CycleGAN Revisit}
CycleGAN introduces two generator-discriminator pairs, $\{G, D_{\mathcal{T}}\}$ and $\{F, D_{\mathcal{S}}\}$, which map a sample from source (target) domain to target (source) domain and produce a sample that is indistinguishable from those in the target (source) domain, respectively.
For generator $G$ and its associated discriminator $D_{\mathcal{T}}$, the adversarial loss is

 \begin{equation}
\begin{split}
\mathcal{L}_{\mathcal{T}adv}(G, D_{\mathcal{T}}, p_{x} , p_{y}) = &\mathbb{E}_{y \sim p_{y}}[(D_{\mathcal{T}}(y)-1)^{2}]\\
&+ \mathbb{E}_{x \sim p_{x}}[(D_{\mathcal{T}}(G(x))^{2}],
\end{split}
\label{Sadv}
\end{equation}where $ p_{x}$ and $ p_{y}$ denote the sample distributions in the source and target domain, respectively. For generator $F$ and its associated discriminator $D_{\mathcal{S}}$, the adversarial loss is
 \begin{equation}
\begin{split}
\mathcal{L}_{\mathcal{S}adv}(F, D_{\mathcal{S}}, p_{y} , p_{x}) = &\mathbb{E}_{x \sim p_{x}}[(D_{\mathcal{S}}(x)-1)^{2}]\\
&+ \mathbb{E}_{y \sim p_{y}}[(D_{\mathcal{S}}(F(y))^{2}].
\end{split}
\label{Tadv}
\end{equation} 

Considering there exist infinitely many alternative mapping functions due to the lack of paired training data, CycleGAN introduces a cycle-consistent loss, which attempts to recover the original image after a cycle of translation and reverse translation, to reduce the space of possible mapping functions. The cycle-consistent loss is
\begin{equation}
\begin{split}
\mathcal{L}_{cyc}(G, F) = &\mathbb{E}_{x \sim p_{x}}[{\Vert F(G(x)) - x\Vert}_{1}]\\
&+ \mathbb{E}_{y \sim p_{y}}[{\Vert G(F(y)) - y\Vert}_{1}].
\end{split}
\label{cycle}
\end{equation}

Apart from cycle-consistent loss and adversarial loss, we use the target domain identity constraint \cite{DBLP:journals/corr/TaigmanPW16} as an auxiliary for image-image translation. Target domain identity constraint is introduced to regularize the generator to be the identity matrix on samples from target domain, written as
\begin{equation}
\begin{split}
\mathcal{L}_{ide}(G, F, p_{x} , p_{y}) = &\mathbb{E}_{x \sim p_{x}}{\Vert F(x) - x \Vert}_{1}\\
&+ \mathbb{E}_{y \sim p_{y}}{\Vert G(y) - y\Vert}_{1}.
\end{split}
\end{equation}
\label{eq:Identity}

As mentioned in \cite{cycle}, generators $G$ and $F$ may change the color of input images without $L_{ide}$. In experiment, we observe that model may generate unreal results without $L_{ide}$ (Fig. \ref{fig3}(b)). 
This is undesirable for re-ID feature learning. 

Thus, we use $L_{ide}$ to preserve the color composition between the input and output (see Section \ref{compar baseline}).
\begin{figure}[t]
\setlength{\abovecaptionskip}{0.2cm}
\setlength{\belowcaptionskip}{-0.5cm}
\begin{center}
\includegraphics[width=1 \linewidth]{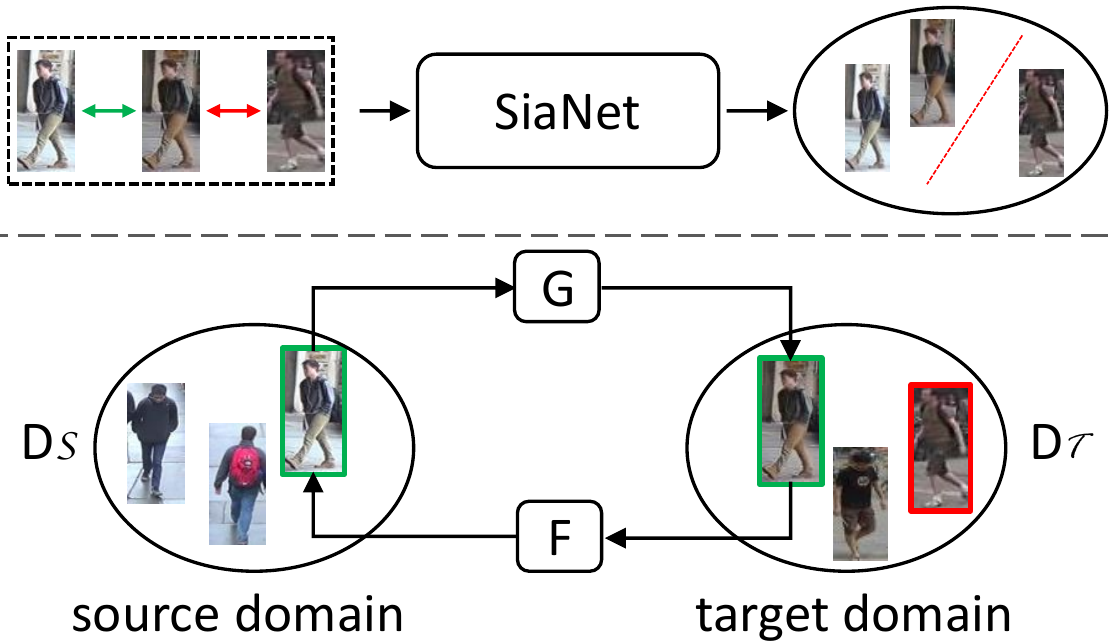}
\end{center}
\caption{SPGAN consists of two components: an  SiaNet (top) and CycleGAN (bottom). CycleGAN learns mapping functions $G$ and $F$ between two domains, and the SiaNet learns a latent space that constrains the learning procedure of mapping functions.}
\label{fig:SPGAN}
\end{figure}

\subsubsection{SPGAN}
Applied in person re-ID, similarity preserving is an essential function to generate improved samples for domain adaptation. As analyzed in Section \ref{Introduction}, we aim to preserve the ID-related information for each translated image. We emphasize that such information should not be the background or image style, but should be underlying and latent. 
To fulfill this goal, we integrate a SiaNet with CycleGAN, as shown in Fig \ref{fig:SPGAN}. During training, CyleGAN is to learn a mapping function between two domains, and SiaNet is to learn a latent space that constrains the learning of mapping function. 

\textbf{Similarity preserving loss function.}
We utilize the contrastive loss \cite{DBLP:conf/cvpr/HadsellCL06} to train SiaNet:
\begin{equation}
\begin{split}
\mathcal{L}_{con}(i, x_{1}, x_{2}) = &(1-i)\{\max(0, m - d)\}^2 + id^2,
\end{split}
\label{eq:Contrastive}
\end{equation}
{where $x_{1}$ and $x_{2}$ are a pair of input vectors, $d$ denotes the Euclidean distance between normalized embeddings of two input vectors, and $i$  represents the binary label of the pair. $i= 1$ if $x_{1}$ and $x_{2}$ are positive pair; $i= 0$ if $x_{1}$ and $x_{2}$ are negative pair.

$m \in [0, 2]$ is the margin that defines the separability in the embedding space.
When $m=0$, the {loss of negative training pair is not back-propagated in the system. When $m > 0$, both positive and negative sample pairs are considered.} A larger $m$ means that the loss of negative training samples has a higher weight in back propagation.

\textbf{Training image pair selection.} In Eq. \ref{eq:Contrastive}, the contrastive loss uses binary labels of input image pairs. The design of the pair similarities reflects the ``self-similarity'' and ``domain-dissimilarity'' principles. Note that, \emph{we select training pairs in an unsupervised manner}, so that we use the contrastive loss without additional annotations.

Formally, CycleGAN has two generators, \ie, generator $G$ which maps source-domain images to the style of the target domain, and generator $F$ which maps target-domain images to the style of the source domain. Suppose two samples denoted as $x_\mathcal{S}$ and $x_\mathcal{T}$ come from the source domain and target domain, respectively. Given $G$ and $F$, we define two positive pairs: 1) $x_{\mathcal{S}}$ and $G(x_\mathcal{S})$, 2) $x_\mathcal{T}$ and $F(x_\mathcal{T})$. In either image pair, the two images contain the same person; the only difference is that they have different styles. In the learning procedure, we encourage the whole network to pull these two images close. 

On the other hand, for generators $G$ and $F$, we also define two types of  negative training pairs: 1) $G(x_\mathcal{S})$ and $x_\mathcal{T}$, 2) $F(x_\mathcal{T})$ and $x_\mathcal{S}$. Such design of negative training pairs is based on the prior knowledge that datasets in different re-ID domains have entirely different sets of IDs. Thus, a translated image should be of  different ID from any target image. In this manner, the network pushes two dissimilar images away. Training pairs are shown in Fig. \ref{fig:carton}. Some positive pairs are also shown in (a) and (d) of each column in Fig. \ref{fig3}.

\textbf{Overall objective function.}
The final SPGAN objective can be written as
\begin{equation}
\begin{split}
\mathcal{L}_{sp}= \mathcal{L}_{\mathcal{T}adv} + \mathcal{L}_{\mathcal{S}adv} + \lambda_{1} \mathcal{L}_{cyc} + \lambda_{2}\mathcal{L}_{ide} + \lambda_{3} \mathcal{L}_{con},
\end{split}
\label{Full Objective}
\end{equation}
\label{eq:Objective}where $\lambda_{t}, t\in \{1,2,3 \}$ controls the relative importance of four objectives. The first three losses belong to the CycleGAN formulation \cite{cycle}, and the contrastive loss induced by SiaNet imposes a new constraint on the system.

\textbf{SPGAN training procedure.} In the training phase, SPGAN are divided into three components which are learned alternately, the generators, discriminators and SiaNet. When the parameters of two components are fixed, the parameters of the third component is updated. We train the SPGAN until the convergence or the maximum iterations. 

\subsection{Feature Learning}\label{sec:feature_learning}
Feature learning is the second step of the ``learning via translation'' framework. Once we have style-transferred dataset $G(\mathcal{S})$ composed of the translated images and their associated labels, the feature learning step is the same as supervised methods. Since we mainly focus on Step 1 (source-target image translation), we adopt the baseline ID-discriminative Embedding (IDE) following the practice in \cite{DBLP:journals/corr/ZhengYH16,zheng2017unlabeled,DBLP:conf/cvpr/ZhongZCL17}. We employ ResNet-50 \cite{DBLP:conf/cvpr/HeZRS16} as the base model and only modify the output dimension of the last fully-connected layer to the number of training identities. During testing, given an input image, we can extract the 2,048-dim Pool5 vector for retrieval under the Euclidean distance.

\captionsetup[subfigure]{labelformat=empty}
\begin{figure}[t]
\setlength{\abovecaptionskip}{-0.2cm} 
\setlength{\belowcaptionskip}{0cm}
	\centering
	\subfloat[  ]{\includegraphics[height=7.5cm]{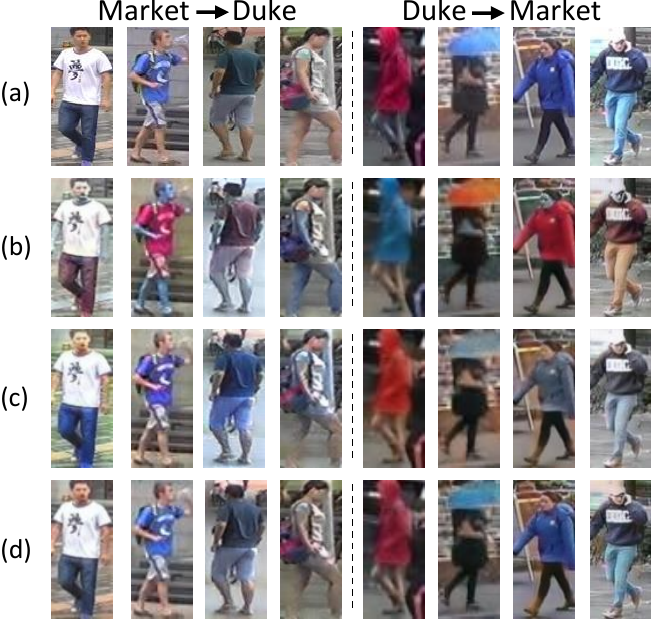}}
	\caption{Visual examples of image-image translation. The left four columns map Market images to the Duke style, and the right four columns map Duke images to the Market style. From top to bottom: (a) original image, (b) output of CycleGAN, (c) output of CycleGAN + $L_{ide}$, and (d) output of SPGAN. Images produced by SPGAN have the target style while preserving the ID information in the source.}
\label{fig3}
\end{figure}
\begin{figure}[t]
\setlength{\abovecaptionskip}{-0.1cm} 
\setlength{\belowcaptionskip}{-0.2cm}
\begin{center}
\includegraphics[width=0.97 \linewidth]{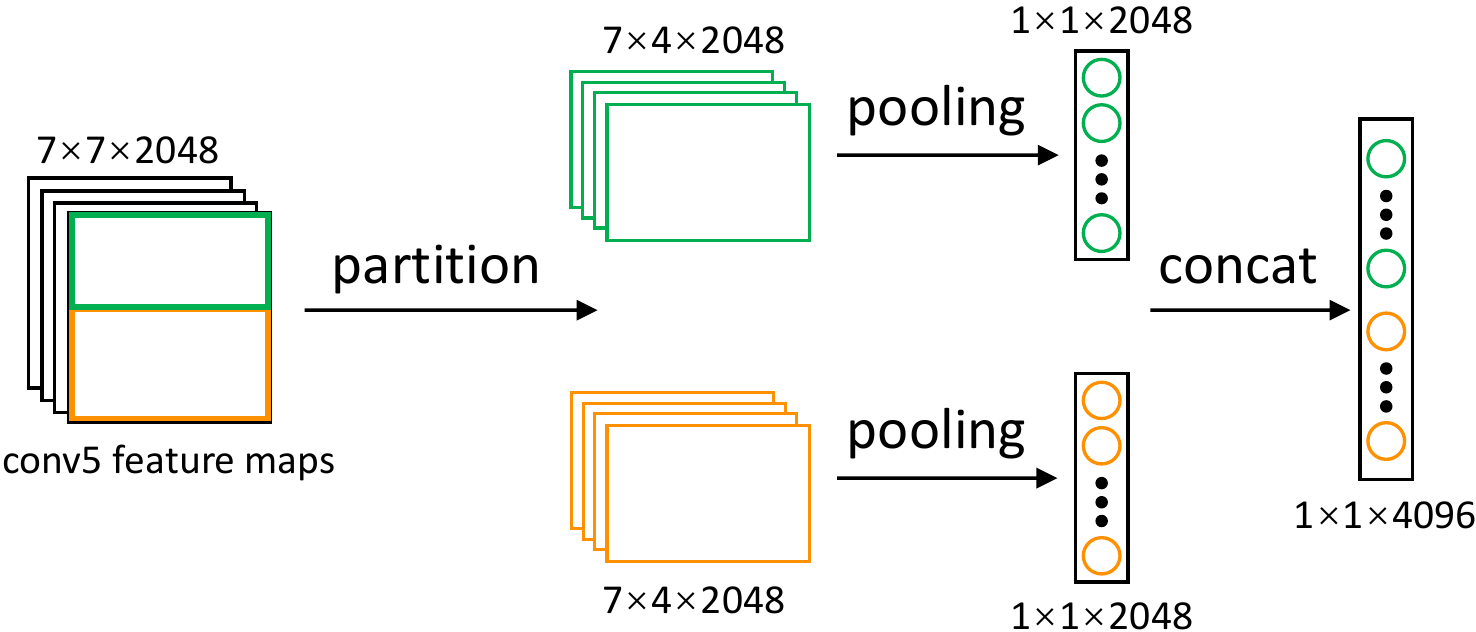}
\end{center}
\caption{Illustration of LMP. We partition the feature map into $P$ $(P=2)$ parts horizontally. We conduct global max/avg pooling on each part and concatenate the feature vectors as the final representation.}
\label{fig:LMP}
\end{figure}
\textbf{Local Max Pooling.} To further improve re-ID performance on the target dataset $\mathcal{T}$, we introduce a feature pooling method named as local max pooling (LMP). It works on a well-trained IDE model and can reduce the impact of noisy signals incurred by the fake translated images. 
In the original ResNet-50, global average pooling (GAP) is conducted on Conv5. In our proposal (Fig. \ref{fig:LMP}), 
we first partition the Conv5 feature maps to $P$ parts horizontally, and then conduct global max/avg pooling on each part. Finally, we concatenate the output of global max pooling (GMP) or GAP of each part as the final feature representation. The procedure is nonparametric, and can be directly used in the testing phase. In the experiment, we will compare local max pooling and local average pooling, and demonstrate the superiority of the former (LMP).

\section{Experiment} \label{experiments}
\subsection{Datasets}
We select two large-scale re-ID datasets for experiment, \ie, \textbf{Market-1501} \cite{DBLP:conf/iccv/ZhengSTWWT15} and \textbf{DukeMTMC-reID} \cite{ristani2016MTMC,zheng2017unlabeled}.  Market-1501  is composed of 1,501 identities, 12,936 training images and 19,732 gallery images (with 2,793 distractors). It is split into 751 identities for training and 750 identities for testing. Each identity is captured by at most 6 cameras. All the bounding boxes are produced by DPM \cite{felzenszwalb2008discriminatively}.  DukeMTMC-reID is a re-ID version of the DukeMTMC dataset \cite{ristani2016MTMC}. 
It contains 34,183 image boxes of 1,404 identities: 702 identities are used for training and the remaining 702 for testing. There are 2,228 queries and 17,661 database images. For both datasets, we adopt rank-1 accuracy and mAP for re-ID evaluation \cite{DBLP:conf/iccv/ZhengSTWWT15}. Sample images of the two datasets are shown in Fig.\ref{fig:sample_images}.
\begin{figure}[t]
\setlength{\abovecaptionskip}{-0.1cm} 
\setlength{\belowcaptionskip}{-0.3cm}
\begin{center}
\includegraphics[width=1 \linewidth]{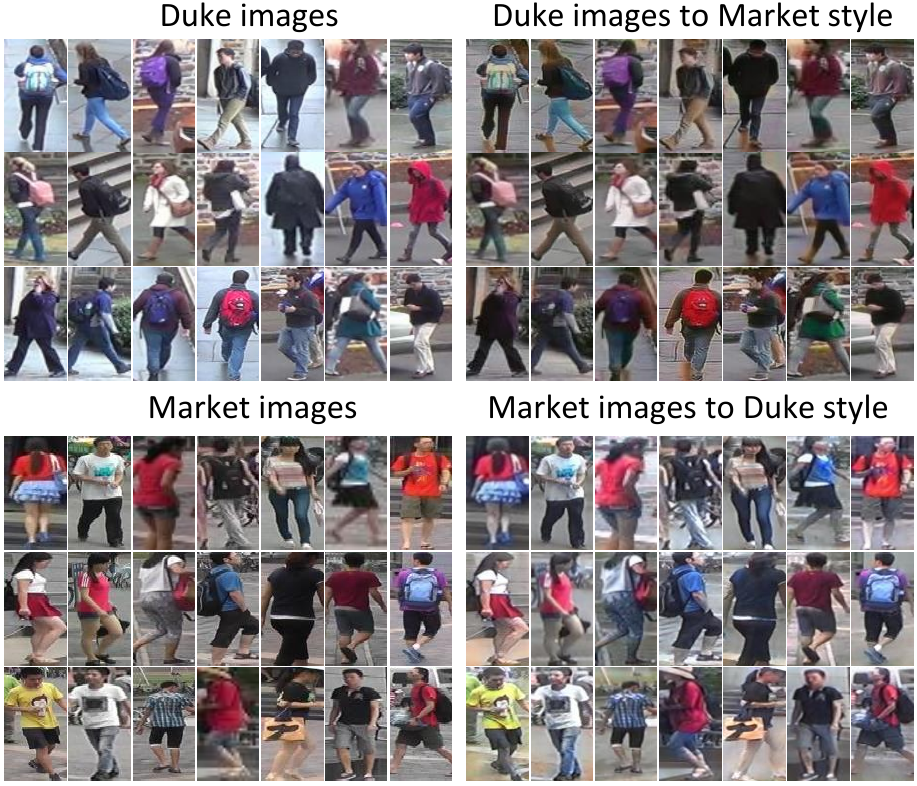}
\end{center}
\caption{Sample images of (upper left:) DukeMTMC-reID dataset, (lower left:) Market-1501 dataset, (upper right:) Duke images which are translated to Market style, and (lower right:) Market images translated to Duke style. We use SPGAN for unpaired image-image translation.}
\label{fig:sample_images}
\end{figure}
\setlength{\tabcolsep}{5pt}
\begin{table*}[t]
\begin{center}
\begin{tabular}{l|ccccc|ccccc}
\hline
\multicolumn{1}{c|}{\multirow{2}{*}{Methods}}&\multicolumn{5}{c|}{DukeMTMC-reID}&\multicolumn{5}{c}{Market-1501}\\
\cline{2-11}
\multicolumn{1}{c|}{}&rank-1&rank-5&rank-10&rank-20&mAP&rank-1&rank-5&rank-10&rank-20&mAP\\
\hline
\hline
Supervised Learning &66.7&79.1 &83.8 & 88.7&46.3&75.8&89.6&92.8&95.4&52.2\\
\hline
Direct Transfer &33.1&49.3&55.6 &61.9&16.7 &43.1&60.8&68.1&74.7&17.0\\
CycleGAN (basel.) &38.1&54.4&60.5&65.9&19.6&45.6 & 63.8&71.3 &77.8&19.1 \\
CycleGAN (basel.) + $L_{ide}$ &38.5&54.6&60.8&66.6&19.9&48.1&66.2&72.7&80.1&20.7\\
\hline
SPGAN ($m=0$) &37.7&53.1&59.5&65.6&20.0&49.2&66.9&74.0&80.0&20.5\\
SPGAN ($m=1$) &39.5&55.0&61.4&67.3&21.0&48.7&65.7&73.0&79.3&21.0\\
SPGAN ($m=2$)&{41.1}&{56.6}&{63.0}&{69.6}&{22.3}&{51.5}&{70.1}&{76.8}&{82.4}&{22.8}\\
{SPGAN ($m=2$) + LMP}&\textbf{46.9}&\textbf{62.6}&\textbf{68.5}&\textbf{74.0}&\textbf{26.4}&\textbf{58.1}&\textbf{76.0}&\textbf{82.7}&\textbf{87.9}&\textbf{26.9} \\
\hline
\end{tabular}
\end{center}
\setlength{\abovecaptionskip}{0cm} 
\setlength{\belowcaptionskip}{-0.1cm} 
\caption{Comparison of various methods on the target domains. When tested on DukeMTMC-reID, Market-1501 is used as source, and vice versa. ``Supervised learning'' denotes using labeled training images on the corresponding target dataset. ``Direct Transfer'' means directly applying the source-trained model on the target domain (see Section \ref{sec:baseline}). By varying $m$ specified in Eq. \ref{eq:Contrastive}, the sensitivity of SPGAN to the relative importance of the positive and negative pairs is shown. When local max pooling (LMP) is applied, the number of parts is set to 7. We use IDE \cite{DBLP:journals/corr/ZhengYH16}  for feature learning.} \label{Compare base}
\label{table:cmpbasl}
\end{table*}

\subsection{Implementation Details} \label{implementation detail}
\textbf{SPGAN training and testing.} We use Tensorflow \cite{tensorflow} to train SPGAN using the training images of Market-1501 and DukeMTMC-reID. Note that, we do not use any ID annotation during training procedure. In all experiments, we empirically set $ \lambda_{1}=10, \lambda_{2}=5, \lambda_{3}=2$ in Eq. \ref{Full Objective} and $m=2$ in Eq. \ref{eq:Contrastive}. With an initial learning rate 0.0002, and model stop training after 5 epochs. During the testing procedure, we employ the Generator $G$ for Market-1501 $\to$ DukeMTMC-reID translation and the Generative $F$ for DukeMTMC-reID $\to$ Market-1501 translation. 
The translated images are used to fine-tune the model trained on source images.

{For CycleGAN, we adopt the architecture released by its authors. For SiaNet, it contains 4 convolutional layers, 4 max pooling layers and 1 fully connected (FC) layer, configured as below. (1) Conv. $4 \times 4$, stride = 2, \#feature maps = 64; (2) Max pooling $2 \times 2$, stride = 2; (3) Conv. $4 \times 4$, stride = 2, \#feature maps = 128; (4) Max pooling $2 \times 2$, stride = 2;  (5) Conv. $4 \times 4$, stride = 2, feature maps = 256; (6) Max pool $2 \times 2$, stride = 2; (7) Conv. $4 \times 4$, stride = 2, \#feature maps = 512; (8) Max pooling $2 \times 2$, stride = 2; (9) FC, output dimension = 128.}

\textbf{Feature learning for re-ID.} As described in Section \ref{sec:feature_learning}, we adopt IDE for feature learning.

Specifically, ResNet-50 \cite{DBLP:conf/cvpr/HeZRS16} pretrained on ImageNet is used for fine-tuning on the translated training set. We modify the output of the last fully-connected layer to 751 and 702 for Market-1501 and DukeMTMC-reID, respectively. 
We use mini-batch SGD to train CNN models on a Tesla K80 GPU. Training parameters such as batch size, maximum number epochs, momentum and gamma are set to 16, 50, 0.9 and 0.1, respectively. The initial learning rate is set as 0.001, and decay to 0.0001 after 40 epochs. 

\subsection{Evaluation} \label{compar baseline}

 \textbf{Comparison between  supervised learning and direct transfer.} The supervised learning method and the direct transfer method are specified in Table \ref{table:summary}. When comparing the two methods in Table \ref{table:cmpbasl}, we can clearly observe a large performance drop when directly using a source-trained model on the target domain. For instance,  
 the ResNet-50 model trained and tested on Market-1501 achieves $75.8\%$ in rank-1 accuracy, but drops to $43.1\%$ when trained on DukeMTMC-reID and tested on Market-1501. A similar drop can be observed when DukeMTMC-reID is used as the target domain, which is consistent with the experiments reported in \cite{fan17unsupervised}. The reason behind the performance drop is the bias of data distributions in different domains.

\textbf{The effectiveness of the ``learning via translation'' baseline using CycleGAN.} In this baseline domain adaptation approach (Section \ref{sec:baseline}), we first translate the label images from the source domain to the target domain and then use the translated images to train re-ID models. As shown in Table \ref{table:cmpbasl}, this baseline framework effectively improves the re-ID performance  in the target dataset. Compared to the direct transfer method, the CycleGAN transfer baseline gains $+2.5\%$ improvements in rank-1 accuracy on  Market-1501. When tested on DukeMTMC-reID, the performance gain is +5.0\% in rank-1 accuracy.
{Through such an image-level domain adaptation method,   effective domain adaptation baselines can be learned.} 

\textbf{The impact of the target domain identity constraint.} 
We conduct experiment to verify the influence of the identity loss on performance in Table \ref{table:cmpbasl}.
We arrive at mixed observations. On the one hand, on DukeMTMC-reID, compared with the CycleGAN baseline, CycleGAN + $L_{ide}$ achieves similar rank-1 accuracy and mAP. On the other hand, on Market-1501, CycleGAN + $L_{ide}$ gains $+2.5\%$ and $1.6\%$ improvement in rank-1 accuracy and mAP, respectively. The reason is that Market-1501 has a larger inter-camera variance. When translating Duke images to the Market style, the translated images may be more prone to translation errors induced by the camera variances. Therefore, the identity loss is more effective when Market is the target domain.

As shown in Fig. \ref{fig3}, this loss prevents CycleGAN from generating strangely colored images.

\textbf{SPGAN effect.} On top of the CycleGAN baseline, we replace CycleGAN with SPGAN ($m=2$). The effectiveness of the proposed similarity preserving constraint can be seen in Table \ref{table:cmpbasl}. Compared with Cycle + $L_{ide}$, on DukeMTMC-reID, the similarity preserving constraint leads to $+2.6\%$ and $+2.4\%$ improvement over CycleGAN + $L_{ide}$ in rank-1 accuracy and mAP, respectively. On Market-1501, the gains are $+3.4\%$ and $2.1\%$. The working mechanism of SPGAN consists in preserving the underlying visual cues associated with the ID labels. The consistent improvement suggests that this working mechanism is critical for generating suitable samples for training in the target domain. Examples of translated images by SPGAN are shown in Fig. \ref{fig:sample_images}.

\begin{figure}[t]
\setlength{\abovecaptionskip}{-0.2cm} 
\setlength{\belowcaptionskip}{-0.3cm}
\begin{center}
\includegraphics[width=0.9 \linewidth]{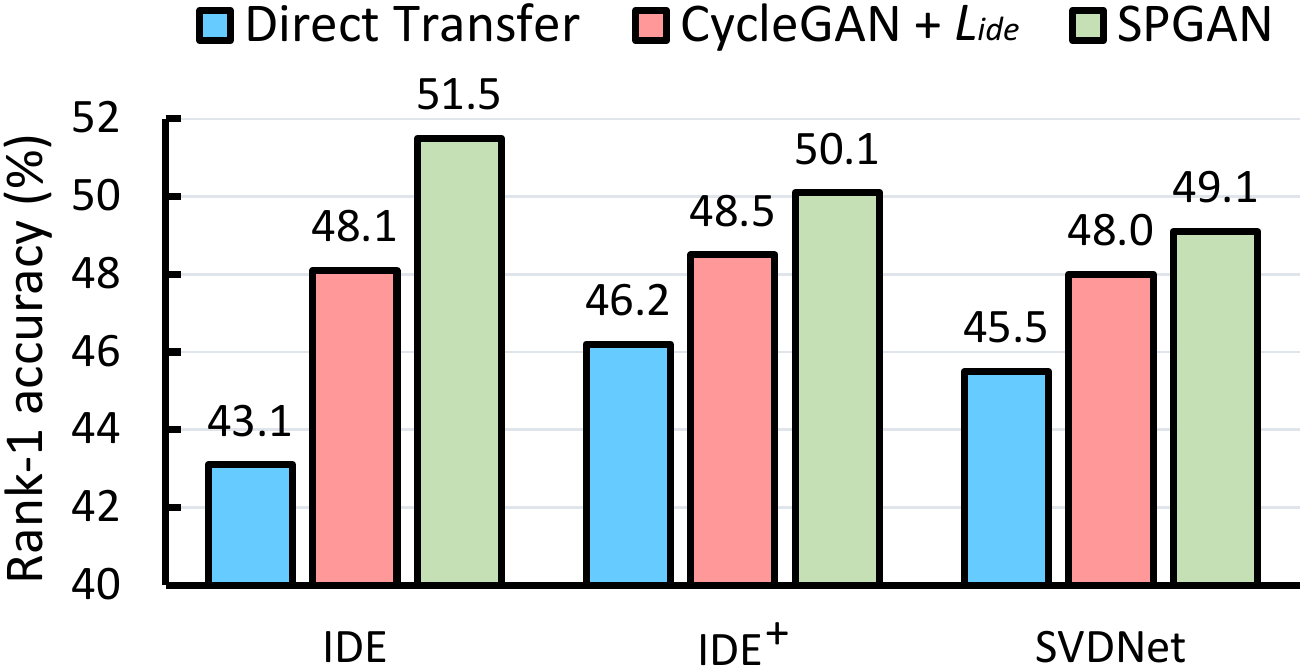}
\end{center}
\caption{Domain adaptation performance with different feature learning methods, including IDE (Section \ref{sec:feature_learning}), IDE$^+$ \cite{zhong2017re}, and SVDNet \cite{SVD}. Three domain adaptation methods are compared, \ie, direct transfer, CycleGAN with identity loss, and the proposed SPGAN. The results are on Market-1501.}
\label{fig:fig6}
\end{figure}

\textbf{Comparison of different feature learning methods.} In Step 2, we evaluate three feature learning methods, \ie, IDE \cite{DBLP:journals/corr/ZhengYH16} (described in Section \ref{sec:feature_learning}), IDE$^+$ \cite{zhong2017re}, and SVDNet \cite{SVD}. Results are shown in Fig. \ref{fig:fig6}. 
An interesting observation is that, while IDE$^+$ and SVDNet are superior to IDE under the scenario of ``Direct Transfer'', the three learning methods are basically on par with each other when using training samples generated by SPGAN.

A possible explanation is that some translated images are noisy, which has a large effect on better learning methods.
\begin{figure}[t]
\setlength{\abovecaptionskip}{-0.2cm} 
\setlength{\belowcaptionskip}{-0.2cm}
\begin{center}
\includegraphics[width=0.75 \linewidth]{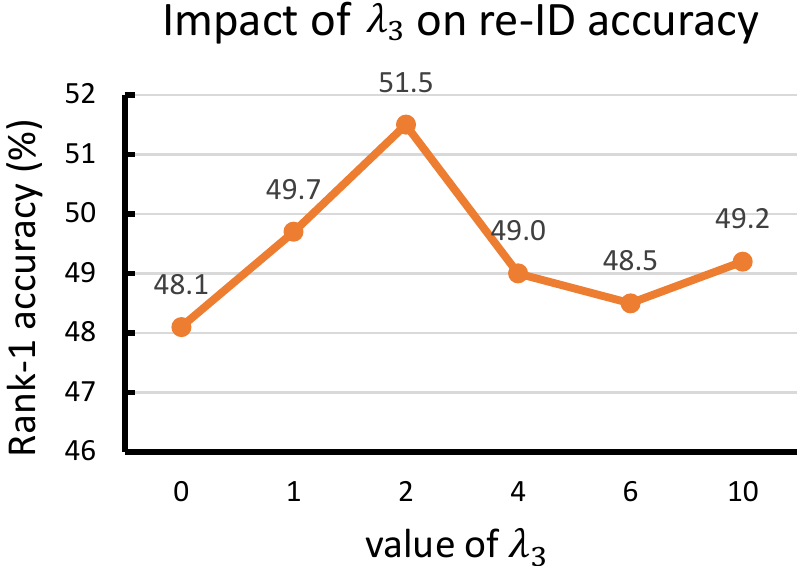}
\end{center}
\caption{$\lambda_{3}$ (Eq. \ref{Full Objective}) \emph{v.s} re-ID accuracy. A larger $\lambda_{3}$ means larger weight of  similarity preserving constraint.}
\label{fig:fig8}
\end{figure}

\textbf{Sensitivity of SPGAN to key parameters.}
The margin $m$ defined in Eq. \ref{eq:Contrastive} is a key parameter. If $m=0$, the loss of negative pairs is not back propagated. If $m$ gets larger, the weight of negative pairs in loss calculation increases.  We conduce experiment to verify the impact of $m$, and results are shown in Table \ref{table:cmpbasl}. When turning off the contribution of negative pairs in Eq. \ref{eq:Contrastive}, ($m=0$), SPGAN only marginally improves the accuracy on Market-1501, and even compromises the system on Duke. When increasing $m$ to 2, we have much superior accuracy. It indicates that the negative pairs are critical to the system. 

Moreover, we evaluate the impact of $\lambda_{3}$ in Eq. \ref{Full Objective} on Market-1501. $\lambda_{3}$ controls the relative importance of the proposed similarity preserving constraint. As shown in Fig. \ref{fig:fig8}, the proposed  constraint is proven effective when compared to $\lambda_{3}=0$, but a larger $\lambda_{3}$ does not bring more gains in accuracy. Specifically, $\lambda_{3}=2$ yields the best accuracy.

\setlength{\tabcolsep}{3pt}
\begin{table}[t]
\setlength{\belowcaptionskip}{-0.7cm}
\begin{center}
\begin{tabular}{l|c|c|cc|cc}
\hline
\multicolumn{1}{l|}{\multirow{2}{*}{\#parts}}&{\multirow{2}{*}{mode}}&{\multirow{2}{*}{dim}}&\multicolumn{2}{c|}{DukeMTMC-reID}&\multicolumn{2}{c}{Market-1501}\\
\cline{4-7}
\multicolumn{1}{c|}{}& & &rank-1&mAP&rank-1&mAP\\
\hline 
\hline
\quad\multirow{2}{*}{1}&Avg&\multirow{2}{*}{2048}&41.1 & 22.3& 51.5&22.8 \\
&Max& & 44.3 &25.0 & 55.7& 21.8 \\
\hline 
\quad{\multirow{2}{*}{2}}&Avg&\multirow{2}{*}{4096}&42.3 & 23.3& 54.4& 25.0\\
&Max& &45.6 & 25.5&57.3 & 26.2\\
\hline 
\quad{\multirow{2}{*}{3}}&Avg&\multirow{2}{*}{6144}& 43.1 & 23.6 & 54.9 & 25.5  \\
&Max& &45.5 & 25.6& 57.4&26.4 \\
\hline 
\quad{\multirow{2}{*}{7}}&Avg&\multirow{2}{*}{14336}&44.2 & 24.4&56.0& 26.1\\
&Max& &\textbf{46.9} &\textbf{26.4} & \textbf{58.1} &\textbf{26.9}   \\
\hline
\end{tabular}
\caption{Performance of various pooling strategies with different numbers  of parts ($P$) and pooling modes (maximum or average) over SPGAN. The best results are in \textbf{bold}.}
\label{table:LMP}
\end{center}
\end{table}

\textbf{Local max pooling.} We apply the LMP on the Conv5 layer to mitigate the influence of noise. Note that LMP is directly adopted in the feature extraction step for testing without fine-tuning. We empirically study how the number of parts and the pooling mode affect the performance. Experiment is conducted on SPGAN. The performance of various numbers of parts ($P = 1, 2, 3, 7$) and different pooling modes (max or average) is provided in Table \ref{table:LMP}. When we use average pooling and $P=1$, we have the original GAP used in ResNet-50. From these results, we speculate that with more parts, a finer partition leads to higher discriminative descriptors and thus higher re-ID accuracy. 

Moreover, we test LMP on supervised learning and domain adaptation scenarios with three feature learning methods, \ie, IDE \cite{DBLP:journals/corr/ZhengYH16}, IDE$^+$ \cite{zhong2017re}, and SVDNet \cite{SVD}. As shown in Fig. \ref{fig:fig8}, LMP does not guarantee stable improvement on supervised learning as observed in ``IDE$^+$'' and SVDNet.

However, when applied in the scenario of domain adaptation, LMP yields improvement over IDE, IDE$^+$, and SVDNet.
The superiority of LMP probably lies in that max pooling filters out some detrimental signals in the descriptor induced by noisy translated images.
\begin{figure}[t]
\setlength{\abovecaptionskip}{-0.3cm} 
\setlength{\belowcaptionskip}{-0.2cm}
\begin{center}
\includegraphics[width=1 \linewidth]{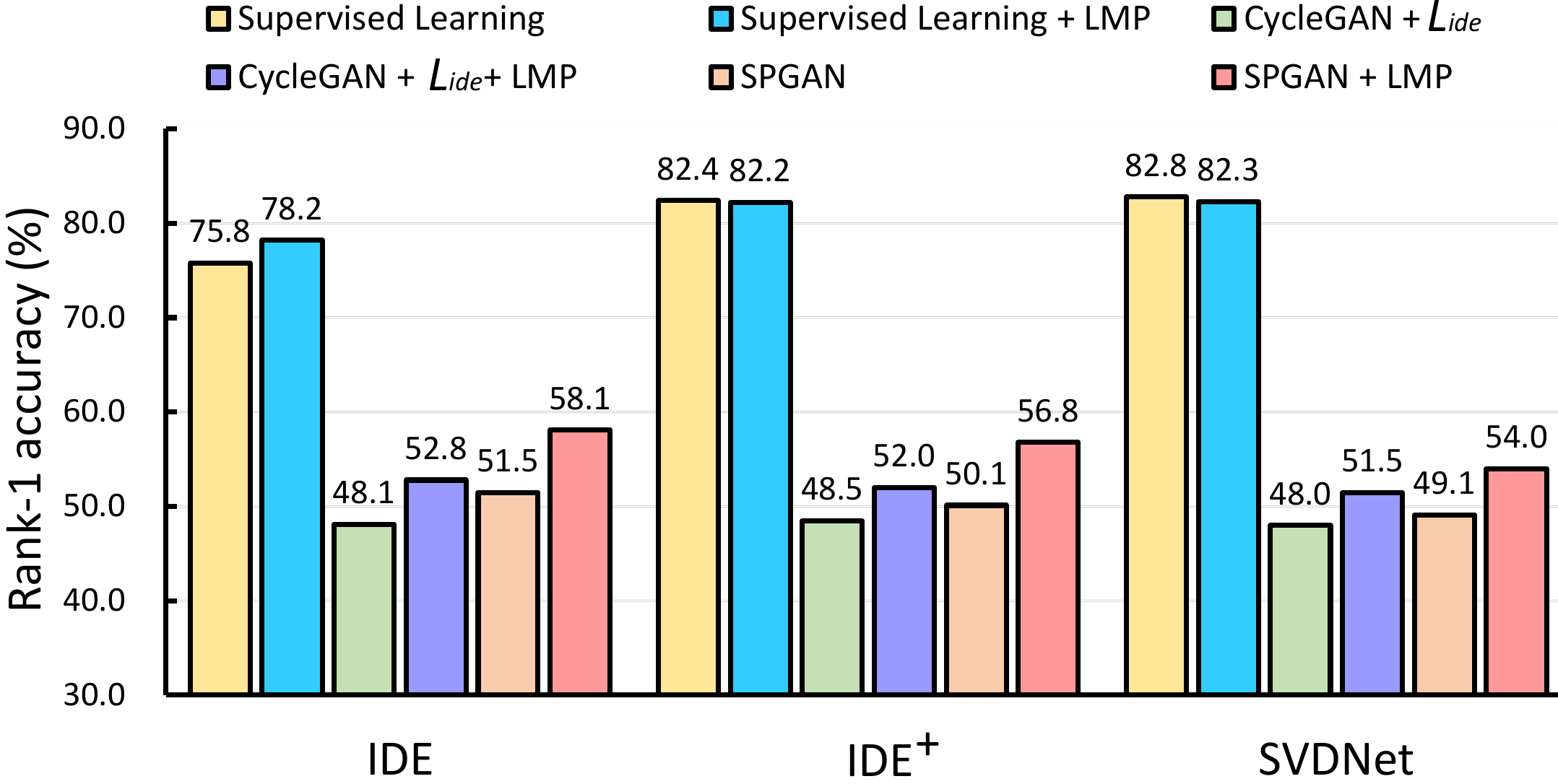}
\end{center}
\caption{Experiment of LMP $(P=7)$ on scenarios of supervised learning and domain adaptation with SPGAN and Cycle + $L_{ide}$. Three feature learning methods are compared, \ie, IDE \cite{DBLP:journals/corr/ZhengYH16}, IDE$^+$ \cite{zhong2017re}, and SVDNet \cite{SVD}. The results are on Market-1501.}
\label{fig:fig8}
\end{figure}

\subsection{Comparison with State-of-the-art Methods }
We compare the proposed method with the state-of-the-art unsupervised learning methods on Market-1501 and DukeMTMC-reID in Table \ref{table:sota-market} and Table \ref{table:sota-duke}, respectively.
\setlength{\tabcolsep}{3pt}
\begin{table}[t]
\setlength{\abovecaptionskip}{-0.2cm} 
\setlength{\belowcaptionskip}{-0.4cm}
\begin{center}
\begin{tabular}{l|c|cccc}
\hline
\multicolumn{1}{l|}{\multirow{2}{*}{Methods}}&\multicolumn{5}{c}{Market-1501}\\
\cline{2-6}
\multicolumn{1}{c|}{}&Setting&Rank-1&Rank-5&Rank-10&mAP\\
\hline
\hline
Bow \cite{DBLP:conf/iccv/ZhengSTWWT15}&SQ&35.8&52.4&60.3&14.8\\
LOMO \cite{DBLP:conf/cvpr/LiaoHZL15}&SQ&27.2&41.6&49.1&8.0\\
\hline
UMDL \cite{DBLP:conf/cvpr/PengXWPGHT16}&SQ&34.5&52.6&59.6&12.4\\
PUL \cite{fan17unsupervised}*&SQ&45.5&60.7&66.7&20.5\\
Direct transfer &SQ&43.1&60.8&68.1&17.0\\
Direct transfer &MQ& 47.9&65.5&73.0&20.6 \\
CAMEL \cite{CAMEL} &MQ&54.5&-&-&26.3\\
\hline
SPGAN&SQ& 51.5&70.1&76.8& 22.8\\
SPGAN&MQ&  {57.0}& {73.9}& {80.3}&  {27.1}\\
{SPGAN+LMP}&SQ&\textbf{58.1}&\textbf{76.0}&\textbf{82.7}&\textbf{26.9}\\
\hline
\end{tabular}
\end{center}
\setlength{\abovecaptionskip}{-0cm} 
\caption{Comparison with state-of-the-art on Market-1501. * denotes unpublished papers. ``SQ'' and ``MQ'' are the single-query and multiple-query settings, respectively. The best results are in \textbf{bold}. }
\label{table:sota-market}
\end{table}

\textbf{Market-1501.} On Market-1501, we first compare our results with two hand-crafted features, \ie, Bag-of-Words (BoW) \cite{DBLP:conf/iccv/ZhengSTWWT15} and local maximal occurrence (LOMO) \cite{DBLP:conf/cvpr/LiaoHZL15}. Those two hand-crafted features are directly applied on test dataset without any training process, their inferiority can be clearly observed.
We also compare existing unsupervised methods, including the Clustering-based Asymmetric MEtric Learning (CAMEL) \cite{CAMEL}, the Progressive Unsupervised Learning (PUL) \cite{fan17unsupervised}, and UMDL \cite{DBLP:conf/cvpr/PengXWPGHT16}. The results of UMDL are reproduced by Fan \emph{et al.} \cite{fan17unsupervised}. In the single-query setting, we achieve rank-1 accuracy = 51.5\% and mAP = 22.8\%. It outperforms the second best method \cite{fan17unsupervised} by +6.0\% in rank-1 accuracy. In the multiple-query setting, we arrive at rank-1 accuracy = 57.0\%, which is +2.5\% higher than CAMEL \cite{CAMEL}. The comparisons indicate the competitiveness of the proposed method on Market-1501.

\textbf{DukeMTMC-reID.} On DukeMTMC-reID, we compare the proposed method with BoW \cite{DBLP:conf/iccv/ZhengSTWWT15}, LOMO \cite{DBLP:conf/cvpr/LiaoHZL15}, UMDL \cite{DBLP:conf/cvpr/PengXWPGHT16}, and PUL \cite{fan17unsupervised} under the single-query setting (there is no multiple-query setting in DukeMTMC-reID). The result obtained by the proposed method is {rank-1 accuracy = 41.1\%, mAP = 22.3\%}. Compared with the second best method, \emph{i.e.,} PUL \cite{fan17unsupervised}, our result is +11.1\% higher in rank-1 accuracy. Therefore, the superiority of SPGAN can be concluded.

\setlength{\tabcolsep}{6pt}
\begin{table}[t]
\setlength{\belowcaptionskip}{-0.2cm}
\begin{center}
\begin{tabular}{l|cccc}
\hline
\multicolumn{1}{l|}{\multirow{2}{*}{Methods}}&\multicolumn{4}{c}{DukeMTMC-reID}\\
\cline{2-5}
\multicolumn{1}{c|}{}&Rank-1&Rank-5&Rank-10&mAP\\
\hline
\hline
Bow \cite{DBLP:conf/iccv/ZhengSTWWT15}&17.1&28.8&34.9&8.3\\
LOMO \cite{DBLP:conf/cvpr/LiaoHZL15}&12.3&21.3&26.6&4.8\\
\hline
UMDL \cite{DBLP:conf/cvpr/PengXWPGHT16}&18.5&31.4&37.6&7.3\\
PUL \cite{fan17unsupervised}*&30.0&43.4&48.5&16.4\\

Direct transfer &33.1&49.3&55.6&16.7\\
\hline
SPGAN& {41.1}& {56.6}& {63.0}& {22.3} \\
{SPGAN+LMP}&\textbf{46.9}&\textbf{62.6}&\textbf{68.5}&\textbf{26.4}\\
\hline

\end{tabular}
\end{center}
\setlength{\abovecaptionskip}{-0.1cm} 
\caption{Comparison with state-of-the-art on DukeMTMC-reID under the single-query setting. * denotes unpublished papers. The best results are in \textbf{bold}. }
\label{table:sota-duke}
\end{table}

\section{Conclusion}
This paper focuses on domain adaptation in person re-ID. When models trained on one dataset are directly transferred to another dataset, the re-ID accuracy drops dramatically due to dataset bias. To achieve improved performance in the new dataset, we present a ``learning via translation'' framework for domain adaptation, characterized by 1) unsupervised image-image translation and 2) supervised feature learning. 
We further propose that the underlying (latent) ID information for the foreground pedestrian should be preserved after image-image translation. To meet this requirement tailored for re-ID, we introduce the unsupervised self-similarity and domain-dissimilarity for similarity preserving image generation (SPGAN). We show that SPGAN better qualifies the generated images for domain adaptation and yields consistent improvement over the CycleGAN. 

\noindent\textbf{Acknowledgment.} Weijian Deng, Qixiang Ye, and Jianbin Jiao are supported by the NSFC under Grant 61671427, 61771447, and Beijing Municipal Science and Technology Commission. Liang Zheng is the recipient of a SIEF STEM+ Business Fellowship, and Yi Yang is the recipient of the Google Faculty Research Award. We thank Pengxu Wei for many helpful comments. 

{\small
\bibliographystyle{ieee}
\bibliography{egbib}
}

\end{document}